\documentclass[letterpaper, 10 pt, conference]{class/ieeeconf}  

\IEEEoverridecommandlockouts                              
\usepackage{stackengine}

\usepackage[sort,nocompress]{cite}

\usepackage{amsmath}
\usepackage{amssymb}
\usepackage{latexsym}
\usepackage{url}
\usepackage{relsize}
\usepackage{multirow}
\usepackage{afterpage}
\usepackage{color}
\usepackage{ifthen}
\usepackage{graphicx}
\usepackage{algpseudocode,algorithm,algorithmicx}
\usepackage{subfigure}
\usepackage{balance}
\usepackage{epstopdf}
\usepackage{gensymb}
\usepackage[bookmarks=false, linkcolor=blue, urlcolor=blue, citecolor=blue]{hyperref} 
\usepackage{booktabs}
\usepackage{makecell}
\usepackage{hhline}
\usepackage{lipsum}
\usepackage{mathtools}

\usepackage{subfloat}
\usepackage{booktabs}
\usepackage{varwidth}
\usepackage{capt-of}

\usepackage[10pt]{moresize}

\hypersetup{pdfstartview={FitH}}

\newcommand\red[1]{{\color{red}#1}}
\newcommand\blue[1]{{\color{blue}#1}}

\title{\LARGE \bf AffordanceNet: An End-to-End Deep Learning Approach \\ for Object Affordance Detection}

\author{Thanh-Toan Do$^{1,\dagger}$, Anh Nguyen$^{2,\dagger}$, Ian Reid$^{1}$%
		\thanks{$^{1}$Thanh-Toan Do and Ian Reid are with the Australian Centre for Robotic Vision (ACRV), University of Adelaide. {\tt \{Thanh-Toan.Do, Ian.Reid\}@adelaide.edu.au}}
        \thanks{$^{2}$Anh Nguyen is with the Department of Advanced Robotics, IIT, Italy. {\tt Anh.Nguyen@iit.it}}        
        \thanks{$^{\dagger}$ Both authors contributed equally to this work.}
}

\begin{document}

\newtheorem{problem}{Problem}
\newtheorem{lemma}{Lemma}
\newtheorem{theorem}[lemma]{Theorem}
\newtheorem{claim}{Claim}
\newtheorem{corollary}[lemma]{Corollary}
\newtheorem{definition}[lemma]{Definition}
\newtheorem{proposition}[lemma]{Proposition}
\newtheorem{remark}[lemma]{Remark}
\newenvironment{LabeledProof}[1]{\noindent{\it Proof of #1: }}{\qed}

\def\beq#1\eeq{\begin{equation}#1\end{equation}}
\def\bea#1\eea{\begin{align}#1\end{align}}
\def\beg#1\eeg{\begin{gather}#1\end{gather}}
\def\beqs#1\eeqs{\begin{equation*}#1\end{equation*}}
\def\beas#1\eeas{\begin{align*}#1\end{align*}}
\def\begs#1\eegs{\begin{gather*}#1\end{gather*}}

\newcommand{\poly}{\mathrm{poly}}
\newcommand{\eps}{\epsilon}
\newcommand{\e}{\epsilon}
\newcommand{\polylog}{\mathrm{polylog}}
\newcommand{\rob}[1]{\left( #1 \right)} 
\newcommand{\sqb}[1]{\left[ #1 \right]} 
\newcommand{\cub}[1]{\left\{ #1 \right\} } 
\newcommand{\rb}[1]{\left( #1 \right)} 
\newcommand{\abs}[1]{\left| #1 \right|} 
\newcommand{\zo}{\{0, 1\}}
\newcommand{\zonzo}{\zo^n \to \zo}
\newcommand{\zokzo}{\zo^k \to \zo}
\newcommand{\zot}{\{0,1,2\}}
\newcommand{\en}[1]{\marginpar{\textbf{#1}}}
\newcommand{\efn}[1]{\footnote{\textbf{#1}}}
\newcommand{\vecbm}[1]{\boldmath{#1}} 
\newcommand{\uvec}[1]{\hat{\vec{#1}}}
\newcommand{\thv}{\vecbm{\theta}}
\newcommand{\junk}[1]{}
\newcommand{\var}{\mathop{\mathrm{var}}}
\newcommand{\rank}{\mathop{\mathrm{rank}}}
\newcommand{\diag}{\mathop{\mathrm{diag}}}
\newcommand{\tr}{\mathop{\mathrm{tr}}}
\newcommand{\acos}{\mathop{\mathrm{acos}}}
\newcommand{\atantwo}{\mathop{\mathrm{atan2}}}
\newcommand{\SVD}{\mathop{\mathrm{SVD}}}
\newcommand{\quadf}{\mathop{\mathrm{q}}}
\newcommand{\linterp}{\mathop{\mathrm{l}}}
\newcommand{\sgn}{\mathop{\mathrm{sign}}}
\newcommand{\sym}{\mathop{\mathrm{sym}}}
\newcommand{\avg}{\mathop{\mathrm{avg}}}
\newcommand{\mean}{\mathop{\mathrm{mean}}}
\newcommand{\erf}{\mathop{\mathrm{erf}}}
\newcommand{\grad}{\nabla}
\newcommand{\R}{\mathbb{R}}
\newcommand{\defeq}{\triangleq}
\newcommand{\dims}[2]{[#1\!\times\!#2]}
\newcommand{\sdims}[2]{\mathsmaller{#1\!\times\!#2}}
\newcommand{\udims}[3]{#1}
\newcommand{\udimst}[4]{#1}
\newcommand{\com}[1]{\rhd\text{\emph{#1}}}
\newcommand{\ind}{\hspace{1em}}
\newcommand{\argmin}[1]{\underset{#1}{\operatorname{argmin}}}
\newcommand{\floor}[1]{\left\lfloor{#1}\right\rfloor}
\newcommand{\step}[1]{\vspace{0.5em}\noindent{#1}}
\newcommand{\quat}[1]{\ensuremath{\mathring{\mathbf{#1}}}}
\newcommand{\norm}[1]{\left\lVert#1\right\rVert}
\newcommand{\ignore}[1]{}
\newcommand{\specialcell}[2][c]{\begin{tabular}[#1]{@{}c@{}}#2\end{tabular}}
\newcommand*\Let[2]{\State #1 $\gets$ #2}
\newcommand{\algorithmicbreak}{\textbf{break}}
\newcommand{\Break}{\State \algorithmicbreak}
\newcommand{\ra}[1]{\renewcommand{\arraystretch}{#1}}

\renewcommand{\vec}[1]{\mathbf{#1}} 

\algdef{S}[FOR]{ForEach}[1]{\algorithmicforeach\ #1\ \algorithmicdo}
\algnewcommand\algorithmicforeach{\textbf{for each}}
\algrenewcommand\algorithmicrequire{\textbf{Require:}}
\algrenewcommand\algorithmicensure{\textbf{Ensure:}}
\algnewcommand\algorithmicinput{\textbf{Input:}}
\algnewcommand\INPUT{\item[\algorithmicinput]}
\algnewcommand\algorithmicoutput{\textbf{Output:}}
\algnewcommand\OUTPUT{\item[\algorithmicoutput]}

\maketitle
\thispagestyle{empty}
\pagestyle{empty}


\begin{abstract}
We propose AffordanceNet, a new deep learning approach to simultaneously detect multiple objects and their affordances from RGB images. Our AffordanceNet has two branches: an object detection branch to localize and classify the object, and an affordance detection branch to assign each pixel in the object to its most probable affordance label. 
The proposed framework employs three key components 
for effectively handling the multiclass problem in the affordance mask:  a sequence of deconvolutional layers, a robust resizing strategy, and a multi-task loss function. 
The experimental results on the public datasets show that our AffordanceNet outperforms recent state-of-the-art methods by a fair margin, while its end-to-end architecture allows 
the inference at the speed of $150ms$ per image. 
This makes our AffordanceNet well suitable for real-time robotic applications. Furthermore, we demonstrate the effectiveness of AffordanceNet in different testing environments and in real robotic applications. The source code is available at \url{https://github.com/nqanh/affordance-net}.

\junk{
 These components play important roles for producing a fine and accurate affordance mask. 
 
 affordance th Our network can be seen as a general design of the recent state-of-the-art instance segmentation frameworks, i.e., our design is to deal with the multiple affordance classes, instead of binary class as in instance segmentation. Different from the recent architectures for instance segmentation, our network contains three novel components: a new loss function, .

We present AffordanceNet, a new deep learning approach to simultaneously detect the object and its affordances from RGB images. Our AffordanceNet has two branches: (1) an object detection branch to localize and classify the object, and (2) an affordance detection branch to assign each pixel in the object to 
 its most probable affordance label. The key contribution of our work is a new method 
 to effectively handle the multiclass problem in the affordance detection task. Furthermore, we show that by deconvoluting the coarse feature map to high resolution, we can achieve smooth and accurate affordance map. 
\blue{The experimental results on the public datasets show that our AffordanceNet outperforms recent state-of-the-art methods by a fair margin, while its end-to-end architecture speeds up the detection, i.e., running at 6 fps. This makes the proposed approach is suitable for real-time robotic applications.}
 Finally, we demonstrate the effectiveness of AffordanceNet in different testing environments, and in real robotic applications. The source code and trained models will be released.
 }
\end{abstract}

\section{INTRODUCTION} \label{Sec:Intro}
An object can be described by various visual properties such as color, shape, or physical attributes such as weight, volume, and material. Those properties are useful to recognize objects or classify them into different categories, however they do not imply the potential actions that human can perform on the object. The capability to understand functional aspects of objects or \textit{object affordances} has been studied for a long time~\cite{Gibson79}. Unlike other visual or physical properties that mainly describe the object alone, affordances indicate functional interactions of object parts with humans. Understanding object affordances is, therefore, crucial to let an autonomous robot interact with the objects and assist humans in various daily tasks.

The problem of modeling object affordances can be considered in different ways. Castellini et al.~\cite{Castellini2011_short} defined affordances in terms of human hand poses during the interaction with objects, while in~\cite{Koppula:2016_short} the authors studied object affordances in the context of human activities. In this work, similar to~\cite{Myers15_short}, we consider object affordances at \textit{pixel} level from an  image, i.e., a group of pixels which shares the same object functionality is considered as one affordance. The advantage of this approach is we can reuse the strong state of the art from the semantic segmentation field, while there is no extra information such as interactions with human is needed. Detecting object affordances, however, is a more difficult task than the classical semantic segmentation problem. For example, two object parts with different appearances may have the same affordance label. It is because the affordance labels are based on the abstract concepts of human actions on the object. Furthermore, it is also essential for an affordance detection method to run in real-time and generalize well on unseen objects.

\begin{figure}[!t] 
\vspace{0.2cm}
    \centering
 	\includegraphics[width=0.99\linewidth, height=0.77\linewidth]{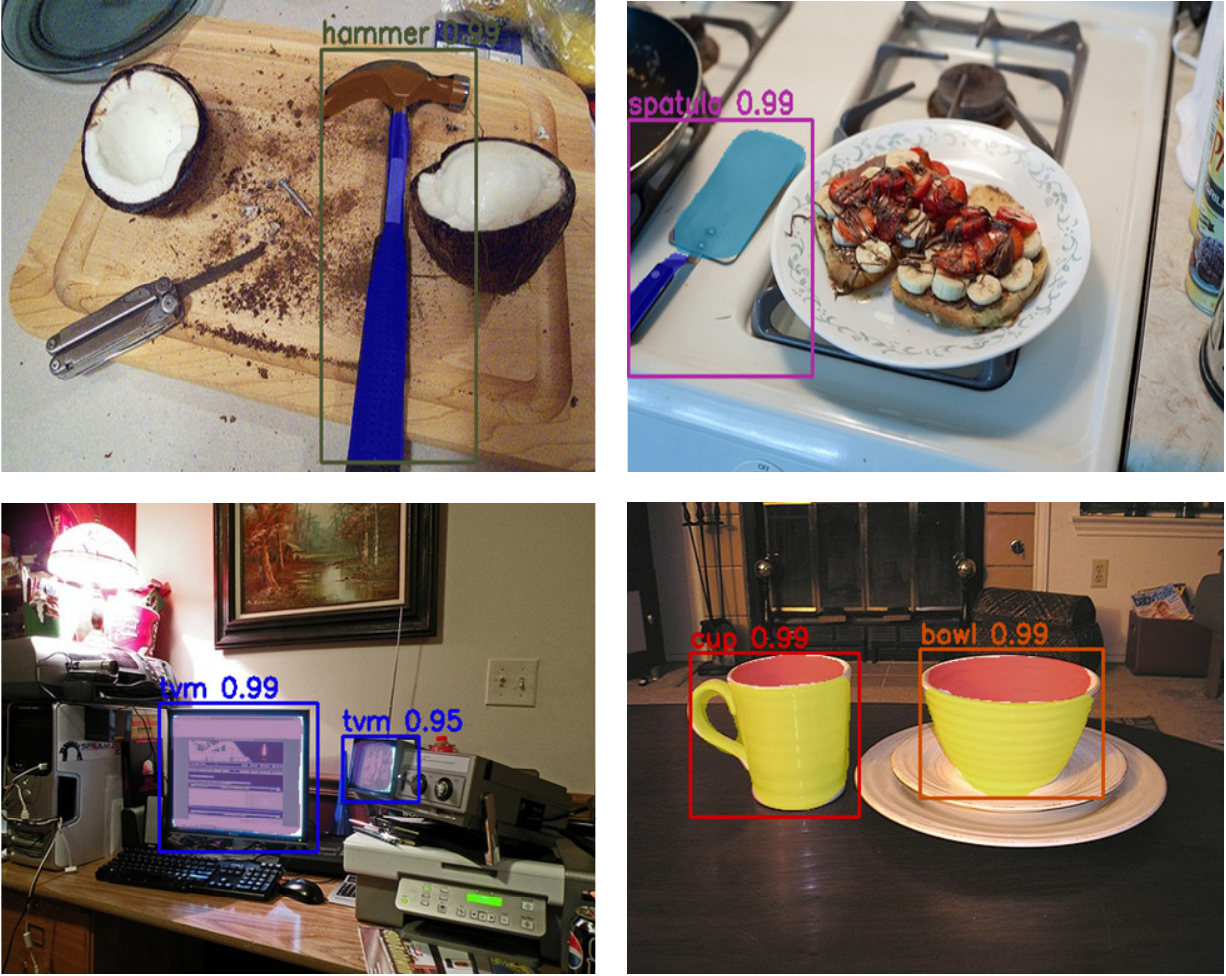} 
    \vspace{1.0ex}
    \caption{Simultaneous affordance and object detection. Some example results of our AffordanceNet, which detects both objects and their multiple affordance classes using an end-to-end architecture.} 
    \label{Fig:intro} 
    
\end{figure}

In many robotic applications, recognizing object affordances is essential, however the robot may still require more information to complete tasks. For example, to pour the water from a \textsl{bottle} into a \textsl{bowl}, the robot not only has to detect object affordances such as \texttt{grasp}, \texttt{contain}, but also be able to localize and recognize the relevant objects (i.e., \textsl{bottle}, \textsl{bowl})~\cite{Nguyen_V2C_ICRA18}. In order to address this, the work in~\cite{Nguyen2017_Aff_short} proposed to use two sequential deep neural networks, one for object detection and one for affordance detection. However, by using two sequential deep networks, it is time consuming during testing, meaning that approach may not be applicable for real-time applications. In this work, we overcome this limitation by using an end-to-end architecture. Our proposed architecture jointly optimizes the object detection and the affordance detection using a multi-task loss function. We show that the proposed method  reduces the complexity during training and testing while improves the overall affordance detection accuracy. Fig~\ref{Fig:intro} shows some example results of our network, which can simultaneously detect the objects and their multiple affordance classes.

In computer vision, simultaneous object detection and object segmentation is  becoming more popular~\cite{Hariharan2014_short}. 
Recent advances in deep learning allow training the detection branch and segmentation branch effectively together. The intuition is that although the detection branch uses object bounding boxes and the segmentation branch uses pixel labels, they can share the same feature maps from the convolutional backbone. The authors in~\cite{Kaiming17_MaskRCNN_short} followed this methodology to build a network for instance segmentation problem and achieved state-of-the-art results. Our work is built upon the works of~\cite{Nguyen2017_Aff_short} and~\cite{Kaiming17_MaskRCNN_short}. However, we differ from \cite{Nguyen2017_Aff_short} by using an end-to-end architecture. We also differ from \cite{Kaiming17_MaskRCNN_short} by having new components, i.e., a new loss function, a sequence of deconvolutional layers, and a robust resizing strategy, for handling the problem of multiple affordance classes. We show that these new components are key factors to achieve high affordance detection accuracy. The experimental results on the public datasets show that our AffordanceNet outperforms recent state-of-the-art methods by a fair margin, while its end-to-end architecture allows inference on a test image in just  $150 ms$. We also demonstrate the effectiveness of AffordanceNet in different testing environments, and in real robotic applications.

The remainder of this paper is organized as follows. We review the related work in Section~\ref{Sec:rw}. We then describe our end-to-end architecture for jointly learning object detection and affordance detection in Section~\ref{Sec:acnn}. In Section~\ref{Sec:exp}, we present the extensive experimental results on the public datasets and the robotic demonstration on a full-size humanoid robot WALK-MAN. Finally, we conclude the paper 
in Section~\ref{Sec:con}.

\junk{ 


}

\section{Related Work} \label{Sec:rw}

The problem of understanding affordances at the pixel level has been termed ``object part labelling'' in the computer vision community, while it is more commonly known as ``affordance detection'' in robotics. In computer vision, the concept of affordances is not restricted to objects, but covers a wide range of applications, from understanding human body parts~\cite{Lin:2017:RefineNet} to environment affordances~\cite{Roy2016_short}~\cite{SceneCut18}, while in robotics, researchers focus more on the real-world objects that the robot can interact with~\cite{Myers15_short}. In~\cite{Schoeler2016_short}, the authors used predefined primary tools to infer object functionalities from 3D point clouds. The work in~\cite{Song15} proposed to combine the global object poses with its local appearances to detect grasp affordances. In \cite{Hedvig2011}, the authors introduced a method to detect object affordances via object-action interactions from human demonstrations. In~\cite{Myers15_short}, the authors used hand-designed geometric features to detect object affordances at pixel level from RGB-D images.

With the rise of deep learning, recent works relied on deep neural networks for designing affordance detection frameworks. The work in~\cite{Lenz14_short} used two deep neural networks to detect grasp affordances from RGB images. The work in~\cite{Nguyen2016_Aff_short} used deep features from Convolutional Neural Networks (CNN) for detecting affordances from RGB-D images. It gained a significant improvement over hand-designed geometric features~\cite{Myers15_short}. 
Similar to~\cite{Nguyen2016_Aff_short}, the work in~\cite{Roy2016_short} introduced multi-scale CNN to localize environment affordances. In~\cite{Sawatzky2017}, to avoid depending on costly pixel groundtruth labels, a weakly supervised deep learning approach was presented to segment object affordances. Recently, in~\cite{Nguyen2017_Aff_short}, the authors proposed to use a deep learning-based object detector to improve the affordance detection accuracy on a real-world dataset. A limitation of that work is that its architecture is not end-to-end -- i.e. two separate networks are used, one for object detection and one for affordance detection -- and this is slow for both training and testing. Furthermore, by training two networks separately, the networks are not jointly optimal. In computer vision, the work of~\cite{Li2016_FCIS} introduced an end-to-end architecture to simultaneously detect and segment object instances. Recently, the authors in~\cite{Kaiming17_MaskRCNN_short} improved over~\cite{Li2016_FCIS} by proposing a region alignment layer which effectively aligns the spatial coordinates of region of interests between the input image space and the feature map space. 

The goal of this work is to simultaneously detect the objects (including the object location and object label) and their associated affordances. We follow the same concept in~\cite{Nguyen2017_Aff_short}, however we use an end-to-end architecture instead of a sequential one. Our object affordance detection network can also be seen as a generalization of the recent state-of-the-art instance segmentation networks~\cite{Kaiming17_MaskRCNN_short}~\cite{Li2016_FCIS}. In particular, our network can detect multiple affordance classes in the object, instead of binary class as in instance segmentation networks~\cite{Kaiming17_MaskRCNN_short}~\cite{Li2016_FCIS}.



\junk{
(e.g. \texttt{walkable}, \texttt{seatable})

}
 
\section{Jointly Affordance and Object Detection} \label{Sec:acnn}

\subsection{Problem Formulation}

Inspired by~\cite{Nguyen2017_Aff_short}~\cite{Kaiming17_MaskRCNN_short}, our framework aims at simultaneously finding the object positions, object classes, and object affordances in images. Follow the standard design in computer vision, the object position is defined by a rectangle with respect to the top-left corner of the image; the object class is defined over the rectangle; the affordances are encoded at every pixel inside the rectangle. The region of pixels on the object that has the same functionality is considered as one affordance. Ideally, we want to detect all relevant objects in the image and map each pixel in these objects to its most probable affordance label.

\begin{figure*}[!t] 
	\vspace{-0.3cm}
    \centering
\includegraphics[scale=0.33]{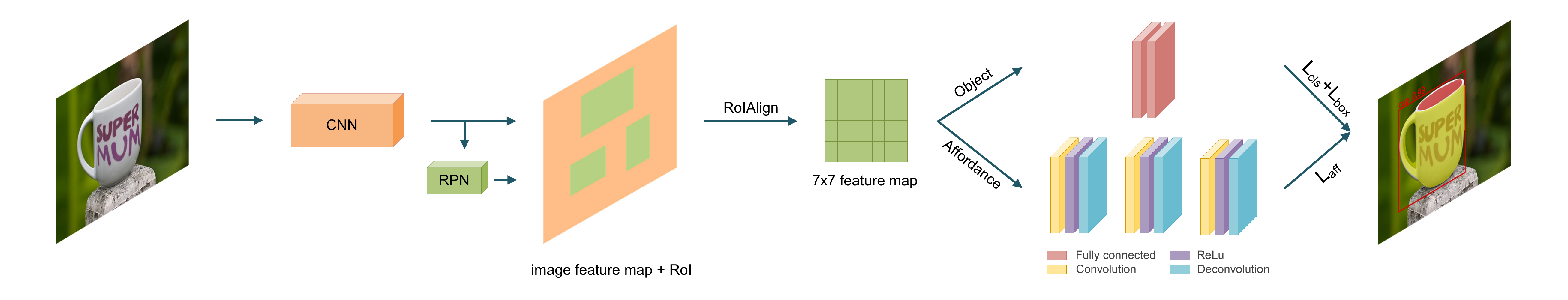}    				
    \vspace{1ex}
    \caption{An overview of our AffordanceNet framework. \textbf{From left to right:} A deep CNN backbone (i.e., VGG) is used to extract image features. The RPN shares weights with the backbone and outputs RoIs. For each RoI, the RoIAlign layer extracts and pools its features (from the image feature map, i.e., the $conv5\_3$ layer of VGG) to a fixed size $7\times 7$ feature map. The object detection branch uses two fully connected layers for regressing object location and classifying object category. The object affordance detection branch consists of a sequence of convolutional-deconvolutional layers and ends  with a softmax layer to output a multiclass affordance mask.}
    \label{fig:overview} 
\end{figure*}

\subsection{AffordanceNet Architecture}

We first describe three main components of our AffordanceNet: the Region of Interest (RoI) alignment layer (RoIAlign)~\cite{Kaiming17_MaskRCNN_short} which is used to correctly compute the feature for an RoI from the image feature map; a sequence of convolution-deconvolution layers to upsample the RoI feature map to high resolution in order to obtain a smooth and fine affordance map; a robust strategy for resizing the training mask to supervise the affordance detection branch. We show that these components are the key factors to achieve high affordance detection accuracy. Finally, we present the whole AffordanceNet architecture in details. Fig.~\ref{fig:overview} shows an overview of our approach.

\subsubsection{RoIAlign}

One of the main components in the recent successful region-based object detectors such as Faster R-CNN~\cite{Ren2015_short} 
is the Region Proposal Network (RPN). This network shares weights with the main convolutional backbone and outputs bounding boxes (RoI / object proposal) at various sizes. For each RoI, a fixed-size small feature map (e.g.,  $7\times7$) is pooled from the image feature map using the RoIPool layer~\cite{Ren2015_short}. The RoiPool layer works by dividing the RoI into a regular grid and then max-pooling the feature map values in each grid cell. This quantization, however, causes misalignments between the RoI and the extracted features due to the harsh rounding operations when mapping the RoI coordinates from the input image space to the image feature map space and when dividing the RoI into grid cells.

In order to address this problem, the authors in~\cite{Kaiming17_MaskRCNN_short} introduced the RoIAlign layer which properly aligns the extracted features with the RoI. Instead of using the rounding operation, the RoIAlign layer uses bilinear interpolation to compute the interpolated values of the input features at four regularly sampled locations in each RoI bin, and aggregates the result using max operation. This alignment technique plays an important role in tasks based on pixel level such as image segmentation.  
We refer the readers to~\cite{Kaiming17_MaskRCNN_short} for a detailed analysis of the RoIAlign  layer.

\subsubsection{Deconvolution for High Resolution Affordance Mask}
In recent state-of-the-art instance segmentation methods such Mask-RCNN~\cite{Kaiming17_MaskRCNN_short} and FCIS~\cite{Li2016_FCIS}, the authors used a small fixed size mask (e.g. $14 \times 14$ or $28 \times 28$) to represent the object segmentation mask. This is feasible since the pixel value in each predicted mask of RoI is binary, i.e., either foreground or background. We empirically found that using small mask size does not work well in the affordance detection problem since we have multiple affordance classes in each object. Hence, we propose to use a sequence of deconvolutional layers for achieving a high resolution affordance mask.

Formally, given an input feature map with size $S_i$, the deconvolutional layer performs the opposite operation of the convolutional layer to create a bigger output map with size $S_o$, in which $S_i$ and $S_o$ are related by:
\begin{equation}
{S_o} = s*({S_i} - 1) + {S_f} - 2*d
\end{equation}
where $S_f$ is the filter size; $s$ and $d$ are stride and padding parameters, respectively.
\begin{figure}[!ht] 
    \centering
	\includegraphics[width=0.95\linewidth, height=0.35\linewidth]{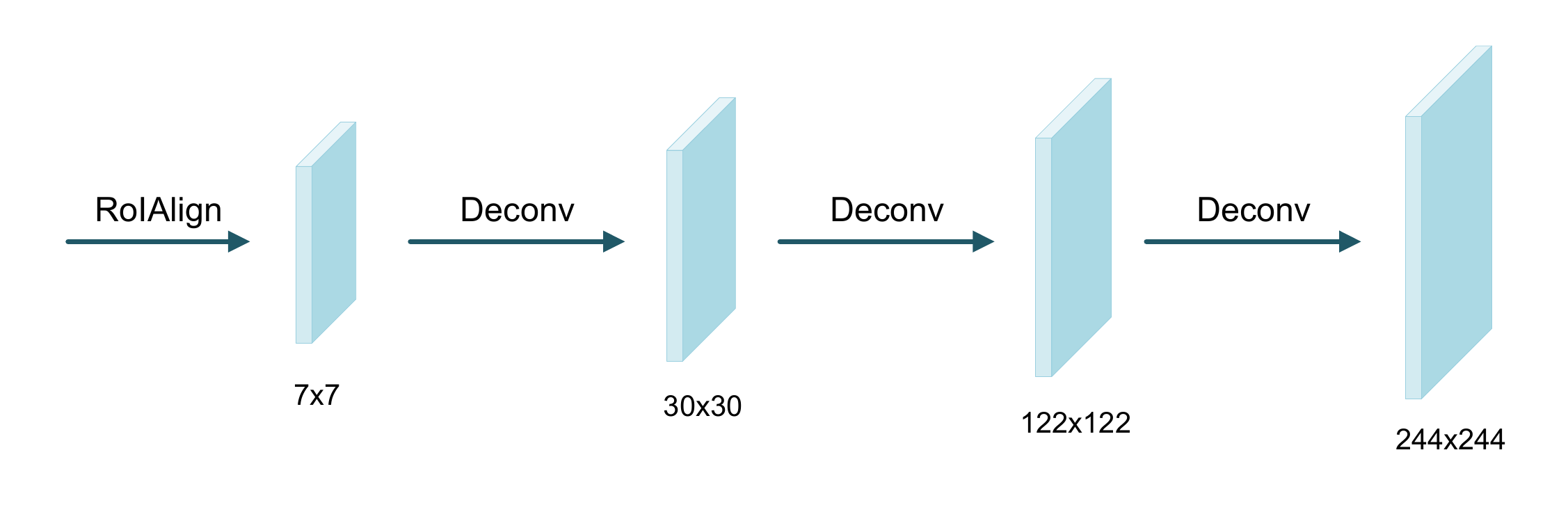}    				
    \vspace{1.5ex}
    \caption{A sequence of three deconvolutonal layers to gradually upsample a $7 \times 7$ fixed size feature map to $244 \times 244$.}
    \label{fig:three_deconvo} 
\end{figure}

In practice, the RoIAlign layer outputs a feature map with size $7 \times 7$. We use three deconvolutional layers to upsample this map to higher resolution (see Fig.~\ref{fig:three_deconvo}). The first deconvolutional layer has the padding $d=1$, stride $s=4$, and kernel size $S_f=8$ to create the map with size $30 \times 30$. Similarly, the second layer has the parameters ($d=1$, $s=4$, $S_f=8$), and the third one has ($d=1$, $s=2$, $S_f=4$) to create the final high resolution map with the size of $244 \times 244$. It is worth noting that before each deconvolutional layer, a convolutional layer (together with ReLu) is used to learn features which will be used for the deconvolution. This convolutional layer can be seen as an adaptation between two consecutive deconvolutional layers. We analyze the effect of the affordance map size in Section~\ref{Sec:exp_effect_mask_size}.

\junk{
The aforementioned strategy can effectively resize the predicted map to the final object affordance map. However, it can not help when the predicted map is too coarse. To address this problem, we propose to deconvolute the feature map to high resolution. We experimentally show that our resize strategy and deconvololuting the feature map are the two key steps to effectively handle the multiple classes in object affordances. 

Formally, given an input feature map $S_i$, the deconvolutional layer performs the opposite operation of the convolutional layer to create the bigger output map $S_o$ as follows:

\begin{equation}
{S_o} = \delta*({S_i} - 1) + {S_f} - 2*\rho
\end{equation}
where $S_f$ is the filter size, $\delta$ and $\rho$ are stride and padding parameters, respectively.

\begin{figure}[!ht] 
    \centering
	\includegraphics[scale=0.38]{figures/3_detection/deconvo.pdf}    				
    \vspace{1.5ex}
    \caption{Lorem ipsum dolor sit amet, consectetuer adipiscing elit. Ut purus elit, vestibulum ut, placerat ac, adipiscing vitae, felis. Lorem ipsum dolor sit amet, consectetuer adipiscing elit. Ut purus elit, vestibulum ut, placerat ac, adipiscing vitae, felis}
    \label{fig:three_deconvo} 
\end{figure}

In practice, the RoIAlign layer outputs a feature map with size $7 \times 7$. We use three deconvolutional layers to upsampling this map to high resolution (see Fig.~\ref{fig:three_deconvo}). The first deconvolutional layer has the padding $\rho=1$, stride $\delta=4$, and kernel size $S_f=8$ to create the map with size $30 \times 30$. Similarly, the second layer has the parameters ($\rho=1$, $\delta=4$, $S_f=8$), and the third one has ($\rho=1$, $\delta=2$, $S_f=4$) to create the high resolution map with the size of $244 \times 244$. We notice that before each deconvolutional layer, we use one convolutional layer with a ReLu layer to gradually learn and upsample the feature map to high resolution. We analyze the effect of the affordance map size in Section~\ref{Sec:exp_effect_mask_size}.
}

\subsubsection{Robust Resizing Affordance Mask}
\label{subsub:resizing}
Similar to Mask-RCNN~\cite{Kaiming17_MaskRCNN_short} and FCIS~\cite{Li2016_FCIS}, our affordance detection branch requires a fixed size (e.g., $244\times 244$) target affordance mask to supervise the training. During training, the authors in~\cite{Kaiming17_MaskRCNN_short}~\cite{Li2016_FCIS} resized the original groundtruth mask of each RoI to the pre-defined mask size to compute the loss. This resizing step outputs a mask with values ranging from 0 to 1, which is thresholded (e.g., at $0.4$) to determine if a pixel is background or foreground. However, using single  threshold value does not work in our affordance detection problem since we have multiple affordance classes in each object.  To address this problem, we propose a resizing strategy with multi-thresholding. 
Given an original groundtruth mask, without loss of generality, let $P=\{c_0,c_1,...,c_{n-1}\}$ be set of $n$ unique labels in that mask, we first linearly map the values in $P$ to $\hat{P}=\{0, 1, ..., n-1\}$ and convert the original mask to a new mask using the mapping from $P$ to $\hat{P}$. We then resize the converted mask to the pre-defined mask size and use the thresholding on the resized mask as follows: 

\begin{equation}\label{Eq_resize_map}
   \rho(x, y)=
    \begin{cases}
      \hat{p}, & \text{if}\ \hat{p} - \alpha  \le \rho(x, y) \le \hat{p} + \alpha \\
      0, & \text{otherwise}
    \end{cases}
\end{equation}
where $\rho(x, y)$ is a pixel value in the resized mask; $\hat{p}$ is one of values in $\hat{P}$; $\alpha$ is the hyperparameter and is set to $0.005$ in our experiments.

Finally, we re-map the values in the thresholded mask back to the original label values (by using the mapping from $\hat{P}$ to $P$) to achieve the target training mask. Note that there is another way to achieve the fixed size target training mask.  We can apply the resizing for each affordance label in the original groundtruth mask separately, i.e., when considering a label, that label is treated as foreground and other labels are treated as background. Then, we can combine the multiple resized masks to achieve the target training mask. However, from the practical point of view, this strategy is time consuming due to the multiple resizing for affordance classes in RoI. 

\junk{As the resizing process outputs a continuous mask, using multi-thresholding directly on the resized mask will generate new affordance labels that do not exist in the original groundtruth mask. In order to address this problem, we propose the following strategy: Given an original groundtruth mask, without loss of generality, let $P=\{c_0,c_1,...,c_{n-1}\}$ be set of $n$ unique labels in the mask, we first linearly map the values in $P$ to $\hat{P}=\{0, 1, ..., n-1\}$, we then convert the original mask to a new mask using the mapping from $P$ to $\hat{P}$, and resize the converted mask to the pre-defined mask size. Note that after resizing, pixel values in the resized mask belong to $[0,n-1]$. Then, we apply the thresholding on the resized mask as follows
\begin{equation}\label{Eq_resize_map}
   \rho(x, y)=
    \begin{cases}
      \hat{p}, & \text{if}\ \hat{p} - \alpha  \le \rho(x, y) \le \hat{p} + \alpha \\
      0, & \text{otherwise}
    \end{cases}
\end{equation}
where $\rho(x, y)$ is a pixel value in the resized mask, $\hat{p}$ is one of values in $\hat{P}$, $\alpha$ is the hyperparameter and is set to $0.005$ in our experiments.

Finally, we re-map the values in the thresholded mask back to the original label values (by using the mapping from $\hat{P}$ to $P$) to achieve the target training mask. Using aforementioned resizing strategy, we avoid generating new affordance labels that do not exist in the original mask. Furthermore, the noisy generated pixels can be considered as background by using a very small threshold ($\alpha=0.005$). Fig.~\ref{fig_resize_map} shows an example of a noisy resized mask by directly resizing the original groundtruth mask and a clean resized mask using our proposed resizing strategy.

\begin{figure}[ht]
  \centering
         \subfigure[]{\label{fig_resize_map_a}\includegraphics[width=2.8cm, height=2.35cm]{figures/3_detection/resize_map/2376_org.jpg}}
    \subfigure[]{\label{fig_resize_map_b}\includegraphics[width=2.8cm, height=2.35cm]{figures/3_detection/resize_map/2376_bad.png}}
     \subfigure[]{\label{fig_resize_map_c}\includegraphics[width=2.8cm, height=2.35cm]{figures/3_detection/resize_map/2376_good.png}}
     
 \vspace{2.5ex}
 \caption{
 The illustration of resized affordance masks using different resizing strategies. \textbf{(a)} The original image. \textbf{(b)} The noisy target mask by directly resizing the original groundtruth mask. \textbf{(c)} The clean target mask by our resizing strategy. Note that the affordance mask in (b) is noisy and contains other labels that do not exist in the original groundtruth mask (best view in color at high resolution).}
 \label{fig_resize_map}
\end{figure}
}

\junk{
Recent state-of-the-art instance segmentation methods such as Mask-RCNN~\cite{Kaiming17_MaskRCNN_short} and FCIS~\cite{Li2016_FCIS} used a relatively small feature map ($14 \times 14$ or $28 \times 28$) to represent the object. This is feasible since the predicted map contains only two classes, i.e., foreground and background. \red{During the training and the testing, this binary map is resized to the original object size. INCORRECT}, hence creating an image with pixel values ranging from $0$ to $1$. 

\red{During training, they resized the groundtruth segmentation mask to $14 \times 14$. During inferencing, they resize the output segmentation mask (i.e., $14 \times 14$) back to the object size. In both resizing stages, they use a thresholding to determine that a pixel is foreground (1) or background (0)}.

These floating values can be converted back to $0$ or $1$ using a simple thresholding heuristic. \red{In general, the final affordance map is achieved by resizing the predicted map to the object size as follows:}

\red{WE ARE TALKING about the MASKRCNN and FCIS, so need a transition to tune to ours.} 

 \begin{equation}\label{Eq_resize_map}
   \rho(x, y)=
    \begin{cases}
      c, & \text{if}\ c - \alpha  \le c \le c + \alpha \\
      0, & \text{otherwise}
    \end{cases}
  \end{equation}
where $\rho$ is a pixel in the final affordance map, $c$ is one of the predicted affordance classes, and $\alpha$ is the threshold.

In the instance segmentation problem, the predicted class is only either foreground or background. The final instance map therefore can easily be achieved by choosing the threshold $\alpha$. For example, the authors in~\cite{Li2016_FCIS} considered pixel value which is smaller than $\alpha=0.4$ as the background, otherwise as the foreground. This strategy, however, does not perform well in our problem since we have multiclass in each object, and the resizing step will generate other affordance labels that can not be removed from the final map by just using the threshold $\alpha$. To address this problem, we propose the following strategy: Given a predicted map with $n$ unique labels $P_m=\{c_0,c_1,...,c_{n-1}\}$, we first linearly map the value in $P_m$ to $\hat{P}_m=\{0, 1, ..., n-1\}$, then resize the $\hat{P}_m$ to the object size. The floating based values in $\hat{P}_m$ are then rounded using Equation~\ref{Eq_resize_map} with $\alpha=0.005$. Finally, we re-map $\hat{P}_m$ to $P_m$ to achieve the final affordance map with the original label set. In this way, we avoid generating other affordance labels that do not exist in the predicted map, while the other noisy generated pixels can be considered as background by using a very small threshold ($\alpha=0.005$). Fig.~\ref{fig_resize_map} shows some examples of the noisy predicted maps and the clean map using our proposed strategy. 
}

\subsubsection{End-to-End Architecture}
Fig.~\ref{fig:overview} shows an overview of our end-to-end affordance detection network. The network is composed of two branches for object detection and affordance detection. 
Given an input image, we use the VGG16~\cite{SimonyanZ14} network as the backbone to extract deep features from the image. A RPN that shares the weights with the convolutional backbone is then used to generate candidate bounding boxes (RoIs). For each RoI, the RoIAlign layer extracts and pools its corresponding features (from the image feature map --- the $conv5\_3$ layer of VGG16) into a $7 \times 7$ feature map. In the object detection branch, we use two fully connected layers, each with $4096$ neurons, followed by a classification layer to classify the object and a regression layer to regress the object location. In the affordance detection branch, the $7 \times 7$ feature map is gradually upsampled to $244 \times 244$ to achieve high resolution map. The affordance branch uses a softmax layer to assign each pixel in the $244 \times 244$ map to its most probable affordance class. The whole network is trained end-to-end using a multi-task loss function.

\junk{Fig.~\ref{fig:overview} shows an overview of our end-to-end affordance detection network. The network is composed of two branches for object detection and affordance detection. Given an input image, we first use a CNN to extract deep features from the image, a RPN that shares  weights with the convolutional backbone is then used to generate candidate bounding boxes. The feature map of each  bounding box is extracted and pooled to a fix-sized feature map by the RoIAlign layer. The object detection branch with two fully connected layers uses the pooled feature maps to localize and classify each object. In the affordance detection branch, we deconvolve the pooled feature map to high resolution using three deconvolutional layers. Before each deconvolutional layer, one convolutional layer and one ReLu layer are used. 

In practice, we use the VGG16~\cite{SimonyanZ14} network as our CNN backbone for simplicity. The RPN shares the weight with the VGG16 network and outputs a set of proposal bounding boxes (RoIs). For each RoI, the RoIAlign layer then extracts and pools its corresponding features (from the image feature map) into a $7 \times 7$ feature map. In the object detection branch, we use two fully connected layers with $4096$ neurons, following by a regression layer to regress the object location and a classification to classify the object. In the affordance detection branch, the $7 \times 7$ feature map is gradually upsampling to $244 \times 244$ to achieve high resolution affordance map. The affordance branch uses a softmax layer to assign each pixel to its most probable affordance class. The whole network is trained end-to-end using a multi-task loss function.}

\subsection{Multi-Task Loss}
In our aforementioned end-to-end architecture, the classification layer outputs a probability distribution $p = (p_0,...,p_K)$ over $K+1$ object categories, including the background. As in~\cite{Ren2015_short}, $p$ is the output of a softmax layer. The regression layer outputs $K+1$ bounding box regression offsets (each offset includes box center and box size): $t^k = (t^k_x,t^k_y,t^k_w,t^k_h)$. Each offset $t^k$ corresponds to each class $k$. Similar to~\cite{Girshick2014}~\cite{Ren2015_short} we parameterize for $t^k$, in which $t^k$ specifies a scale-invariant translation and log-space height/width shift relative to an anchor box of the RPN. The affordance detection branch outputs a set of probability distributions $m = \{m^i\}_{i \in RoI}$ for each pixel $i$ inside the RoI, in which $m^i = (m^i_0,...,m^i_C)$ is the output of a softmax layer defined on $C+1$ affordance labels, including the background. 

We use a multi-task loss $L$ to jointly train the bounding box class, the bounding box position, and the affordance map as follows:
\begin{equation}\label{loss_coarse}
L = {L_{cls}} + {L_{loc}} + {L_{aff}}
\end{equation}
where ${L_{cls}}$ is defined on the output of the classification layer; ${L_{loc}}$ is defined on the output of the regression layer; ${L_{aff}}$ is defined on the output of the  affordance detection branch. 

The prediction target for each RoI is a groundtruth object class $u$, a groundtruth bounding box offset $v$, and a target affordance mask $s$. The values of $u$ and $v$ are provided with the training datasets. The target affordance mask $s$ is the intersection between the RoI and its associated groundtruth mask. For pixels inside the RoI which do not belong to the intersection, we label them as background. Note that the target mask is then resized to a fixed size (i.e., $244\times 244$) using the proposed resizing strategy in Section \ref{subsub:resizing}. 
Specifically, we can rewrite Equation~\ref{loss_coarse} as follows:
\begin{align}\label{eq:loss_fine}
L(p,u,t^u,v,m,s) =& L_{cls}(p,u) + I[u\ge1]L_{loc}(t^u,v) \nonumber \\ 
&+ I[u\ge1]L_{aff}(m,s)
\end{align}
The first loss $L_{cls}(p,u)$ is the multinomial cross entropy loss for the classification and is computed as follows:
\begin{equation}
 L_{cls}(p,u) = -log(p_u)
\end{equation} 
where $p_u$ is the softmax output for the true class $u$. 

The second loss $L_{loc}(t^u,v)$ is \textit{Smooth L1} loss ~\cite{DBLP:conf/iccv/Girshick15} between the regressed box offset $t^u$ (corresponding to the groundtruth object class $u$) and the groundtruth box offset $v$, and is computed as follows:
\begin{equation}
L_{loc}(t^u,v) = \sum_{i\in\{x,y,w,h\}} Smooth_{L1} (t^u_i - v_i)
\end{equation}
where 
\begin{displaymath}
\textrm{$Smooth_{L1} (x)$} = \left\{ \begin{array}{ll}
\textrm{$0.5x^2$} & \textrm{if $|x| <1$}\\
\textrm{$|x-0.5|$} & \textrm{otherwise}
\end{array} \right.
\end{displaymath}

The $L_{aff}(m,s)$ is the multinomial cross entropy loss for the affordance detection branch and is computed as follows:

\begin{equation}
L_{aff}(m,s) = \frac{-1}{N}\sum_{i\in RoI} log(m^i_{s_i})  
\end{equation}
where $m^i_{s_i}$ is the softmax output at pixel $i$ for the true label $s_i$; 
$N$ is the number of pixels  in the RoI.

In Equation (\ref{eq:loss_fine}), $I[u \ge 1]$ is an indicator function which outputs 1 when $u\ge 1$ and $0$ otherwise. This means that we only define the box location loss $L_{loc}$ and the affordance detection loss $L_{aff}$ only on the positive RoIs. While the object classification loss $L_{cls}$ is defined on both positive and negative RoIs. 

It is worth noting that 
our loss for affordance detection branch is different from the instance segmentation loss in~ \cite{Kaiming17_MaskRCNN_short}~\cite{Li2016_FCIS}. In those works, the authors rely on the output of the classification layer to determine the object label. Hence the segmentation in each RoI can be considered as a binary segmentation, i.e., foreground and background. Thus, the authors use per-pixel $sigmoid$ layer and binary cross entropy loss. In our affordance detection problem, the affordance labels are different from the object labels. Furthermore, the number of affordances in each RoI is not binary, i.e., it is always bigger than 2 (including the background). Hence, we rely on a per-pixel $softmax$ and a multinomial cross entropy loss.

\junk{
The detail parameters for deciding positive / negative RoIs outputted from RPN is presented in the next section.  
the state-of-the-art object detection
the standard practice in object detection

 inside the bounding box 
 
 \red{Remember to mention positive and negative boxes in RPN}.

For each proposal bounding box outputted by RPN, our network architecture has three output branches for three losses as follow:

It is worth noting that, follow~\cite{Ren2015_short}, we parameterize $t^k$ with scale-invariant translation and log-space width/height which is relative to the anchor box of RPN.

We use a multi-task loss $L$ on each bounding box to jointly train the bounding box label, the bounding box position, and the affordance mask inside the bounding box:

}

\subsection{Training and Inference}

We train the network in an end-to-end manner using stochastic gradient descent with $0.9$ momentum and $0.0005$ weight decay. The network is trained on a Titan X GPU for $200k$ iterations. The learning rate is set to $0.001$ for the first $150k$ and decreased by $10$ for the last $50k$. The input images are resized such that the shorter edge is $600$ pixels, but the longer edge does not exceed $1000$ pixels. In case the longer edge exceeds $1000$ pixels, the longer edge is set to $1000$ pixels, and the images are resized based on this edge. Similar to~\cite{Kaiming17_MaskRCNN_short}, we use $15$ anchors in the RPN ($5$ scales and $3$ aspect ratios). Top $2000$ RoIs from RPN (with a ratio of 1:3 of positive to negative) are subsequently used for computing the multi-task loss. An RoI is considered positive if it has IoU with a groundtruth box of at least 0.5 and negative otherwise. 

During the inference phase, we select the top $1000$ RoIs produced by the RPN and run the object detection branch on these RoIs, followed by a non-maximum suppression~\cite{Girshick2015_deformable_short}. From the outputs of the detection branch, we select the outputted boxes that have the classification score higher than $0.9$ as the final detected objects. In case there are no boxes satisfying this condition, we select the one with highest classification score as the only detected object. We use the detected objects as the inputs for affordance detection branch. For each pixel in the detected object, the affordance branch predicts $C+1$ affordance classes. The output affordance label  for each pixel  is achieved by taking the maximum across the affordance classes. Finally, the predicted $244 \times 244$ affordance mask of each object is resized to the object (box) size using the resizing strategy in Section~\ref{subsub:resizing}. 
In case there is the overlap between detected objects, similar to \cite{Nguyen2017_Aff_short}, the final affordance label is decided based on the affordance priority.  For example, the ``contain" affordance is considered to have low priority than other affordances since there may have other objects laid on it.

\junk{

As in~\cite{Kaiming17_MaskRCNN_short}, we first run the object detection, then the affordance detection to achieve the affordance map.

} 
\section{EXPERIMENTS} \label{Sec:exp}


\begin{table}[!t]
\centering\ra{1.3}
\caption{Performance on IIT-AFF Dataset}
\renewcommand\tabcolsep{1.9pt}
\label{tb_result_iit}
\hspace{2ex}

\begin{tabular}{@{}rcccccccc@{}}
\toprule 					 &
{\shortstack{\ssmall ED-RGB\\ \ssmall ~\cite{Nguyen2016_Aff_short}} } & 
{\shortstack{\ssmall ED-RGBD\\ \ssmall ~\cite{Nguyen2016_Aff_short}}} & 
{\shortstack{\ssmall DeepLab\\ \ssmall~\cite{Chen2016_deeplab}}} &
{\shortstack{\ssmall DeepLab-\\ \ssmall CRF~\cite{Chen2016_deeplab}}}  & 
{\shortstack{\ssmall BB-CNN\\ \ssmall ~\cite{Nguyen2017_Aff_short}}} &
{\shortstack{\ssmall BB-CNN-\\ \ssmall CRF~\cite{Nguyen2017_Aff_short}}} &
{\shortstack{\ssmall AffordanceNet\\ \ssmall (ours)}} &  \\

\midrule
\texttt{contain} 				& 66.38   & 66.00   & 68.84	& 69.68	& 75.60     & 75.84   & \textbf{79.61} \\
\texttt{cut}					& 60.66   & 60.20   & 55.23	& 56.39	& 69.87     & 71.95   & \textbf{75.68} \\
\texttt{display}				& 55.38   & 55.11   & 61.00	& 62.63 & 72.04     & 73.68   & \textbf{77.81} \\
\texttt{engine} 				& 56.29   & 56.04   & 63.05	& 65.11	& 72.84     & 74.36   & \textbf{77.50} \\
\texttt{grasp}					& 58.96   & 58.59   & 54.31	& 56.24	& 63.72     & 64.26   & \textbf{68.48} \\
\texttt{hit} 					& 60.81   & 60.47   & 58.43	& 60.17	& 66.56     & 67.07   & \textbf{70.75} \\
\texttt{pound} 					& 54.26   & 54.01   & 54.25	& 55.45	& 64.11     & 64.86   & \textbf{69.57} \\
\texttt{support} 				& 55.38   & 55.08   & 54.28	& 55.62	& 65.01     & 66.12   & \textbf{69.81} \\
\texttt{w-grasp} 				& 50.66   & 50.42   & 56.01	& 57.47	& 67.34     & 68.41   & \textbf{70.98} \\
\cline{1-8}
\textbf{Average}				& 57.64   & 57.32   & 58.38	& 59.86	& 68.57     & 69.62   & \textbf{73.35} \\
\bottomrule
\end{tabular}
\end{table}


\subsection{Dataset and Baseline}


\textbf{IIT-AFF Dataset} The IIT-AFF dataset is recently introduced in~\cite{Nguyen2017_Aff_short} and consists of $8,835$ real-world images. This dataset is suitable for deep learning methods and robotic applications since around $60\%$ of the images are from ImageNet dataset~\cite{Russakovsky2015}, while the rest images are taken by the authors from cluttered scenes. In particular, this dataset contains $10$ object categories, $9$ affordance classes, $14,642$ object bounding boxes, and $24,677$ affordance regions at pixel level. We use the standard split as in~\cite{Nguyen2017_Aff_short} to train our network (i.e. $70\%$ for training and $30\%$ for testing).

\textbf{UMD Dataset} The UMD dataset~\cite{Myers15_short} contains around $30,000$ RGB-D images of daily kitchen, workshop, and garden objects. The RGB-D images of this dataset were captured from a Kinect camera on a rotating table in a clutter-free setup. This dataset has $7$ affordance classes and $17$ object categories. Since there is no groundtruth for the object bounding boxes, we compute the rectangle coordinates of object bounding boxes based on the affordance masks. We use only the RGB images of this dataset and follow the split in~\cite{Myers15_short} to train and test our network.

\begin{figure*}[!t]
  \centering
\includegraphics[scale=0.265]{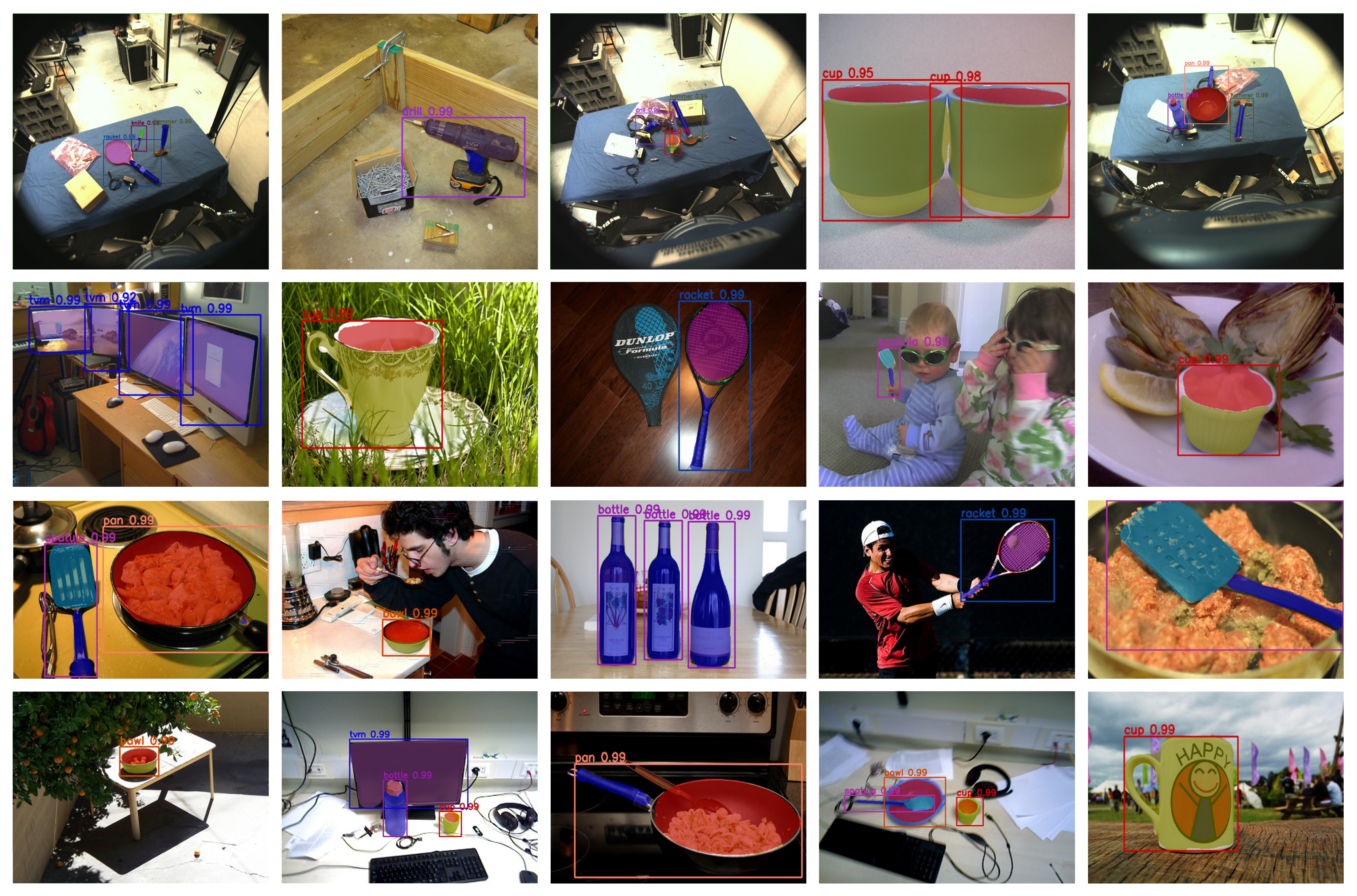}
\vspace{-1ex}
 \caption{Examples of affordance detection results by AffordanceNet on the IIT-AFF dataset.}
 \label{Fig:result_aff_detection}
\end{figure*}

\textbf{Baseline}
As the standard practice, we use the $F_\beta ^w$ metric~\cite{Margolin14_short} to evaluate the affordance detection results. We compare our AffordanceNet with the following state-of-the-art approaches: DeepLab~\cite{Chen2016_deeplab} with and without post processing with CRF (denoted as DeepLab and DeepLab-CRF), CNN with encoder-decoder architecture~\cite{Nguyen2016_Aff_short} on RGB and RGB-D images (denoted as ED-RGB and ED-RGBD), CNN with object detector (BB-CNN) and CRF (BB-CNN-CRF)~\cite{Nguyen2017_Aff_short}. For the UMD dataset, we also report the results from the geometric features-based approach (HMD and SRF)~\cite{Myers15_short} and a deep learning-based approach that used both RGB and depth images as inputs (ED-RGBHHA)~\cite{Nguyen2016_Aff_short}. Note that, all the deep learning-based methods use the VGG16 as the main backbone for a  fair comparison.

\begin{table}[!]
\centering\ra{1.3}
\renewcommand\tabcolsep{2.7pt}
\caption{Performance on UMD Dataset}
\label{tb_result_umd}
\hspace{2ex}
\begin{tabular}{@{}rcccccccc@{}}
\toprule 					
& \shortstack{\ssmall HMP\\ \ssmall ~\cite{Myers15_short}}   
& \shortstack{\ssmall SRF\\ \ssmall ~\cite{Myers15_short}}  
& \shortstack{\ssmall DeepLab\\ \ssmall ~\cite{Chen2016_deeplab}}    
& \shortstack{\ssmall ED-RGB\\ \ssmall ~\cite{Nguyen2016_Aff_short}}    
& \shortstack{\ssmall ED-RGBD\\ \ssmall ~\cite{Nguyen2016_Aff_short}}     
& \shortstack{\ssmall ED-RGB\\ \ssmall HHA~\cite{Nguyen2016_Aff_short}}   
& \shortstack{\ssmall AffordanceNet\\ \ssmall (ours)}   
\\
\midrule
\texttt{grasp} 				& 0.367   & 0.314   & 0.620				& 0.719	    		& 0.714     & 0.673  & \textbf{0.731} \\
\texttt{w-grasp}			& 0.373   & 0.285   & 0.730				& 0.769  		    & 0.767     & 0.652  & \textbf{0.814} \\
\texttt{cut}				& 0.415   & 0.412   & 0.600				& 0.737   			& 0.723     & 0.685  & \textbf{0.762} \\
\texttt{contain}			& 0.810   & 0.635   & \textbf{0.900}	& 0.817   			& 0.819     & 0.716  & 0.833  \\
\texttt{support} 			& 0.643   & 0.429   & 0.600				& 0.780   			& 0.803     & 0.663  & \textbf{0.821} \\
\texttt{scoop} 				& 0.524   & 0.481  	& \textbf{0.800}	& 0.744  			& 0.757     & 0.635  & 0.793  \\
\texttt{pound} 				& 0.767   & 0.666   & \textbf{0.880}	& 0.794   			& 0.806     & 0.701  & 0.836  \\
\cline{1-8}	
\textbf{Average}			& 0.557   & 0.460   & 0.733				& 0.766   			& 0.770     & 0.675  & \textbf{0.799} \\		
\bottomrule
\end{tabular}
\end{table}
\subsection{Results}

\textbf{IIT-AFF Dataset} Table~\ref{tb_result_iit} summarizes results on the IIT-AFF dataset. 
The results clearly show that AffordanceNet significantly improves over the state of the art. In particular, AffordanceNet boosts the $F_\beta ^w$ score to $73.35$, which is  $3.7\%$ improvement over the second best BB-CNN-CRF. It is worth noting that AffordanceNet achieves this result using an end-to-end architecture, and no further post processing step such as CRF is used. Our AffordanceNet also achieves the best results for all $9$ affordance classes. We also found that for the dataset containing cluttered scenes such as IIT-AFF, the approaches that combine the object detectors with deep networks to predict the affordances (AffordanceNet, BB-CNN) significantly outperform over the methods that use deep networks alone (DeepLab, ED-RGB).

\begin{figure*}
\centering
\footnotesize
  \stackunder[7pt]{\includegraphics[width=2.88cm, height=2.45cm]{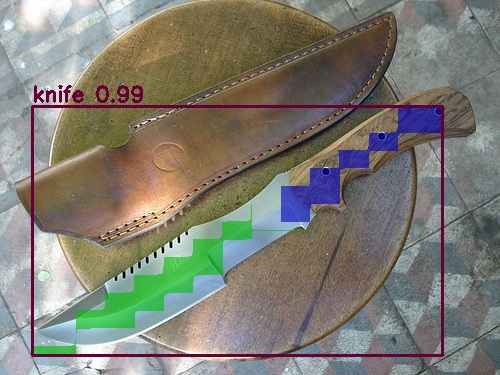}} {AffordanceNet14}
  \stackunder[7pt]{\includegraphics[width=2.88cm, height=2.45cm]{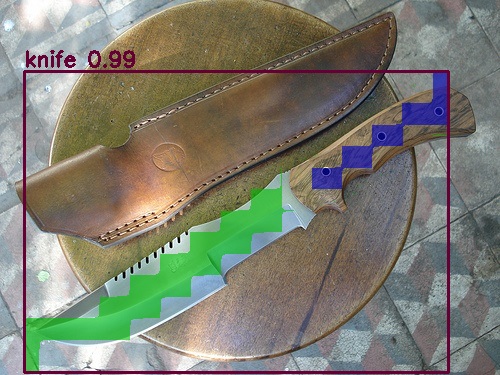}} {AffordanceNet14\_6conv} 
  \stackunder[7pt]{\includegraphics[width=2.88cm, height=2.45cm]{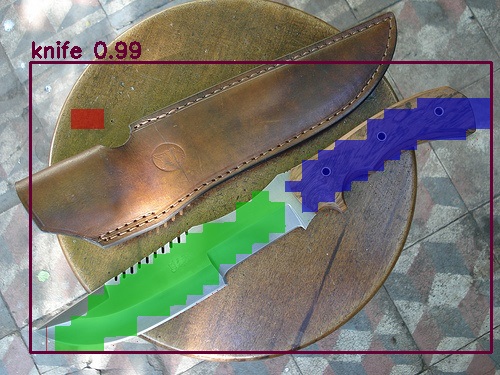}} {AffordanceNet28} 
  \stackunder[7pt]{\includegraphics[width=2.88cm, height=2.45cm]{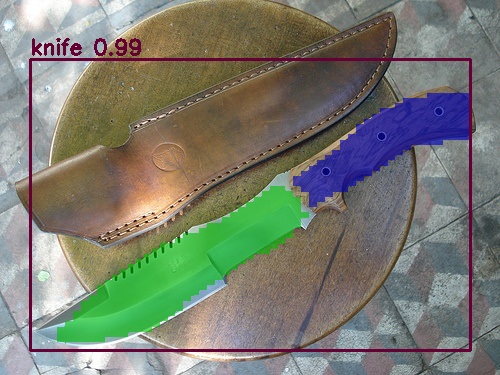}} {AffordanceNet56}
  \stackunder[7pt]{\includegraphics[width=2.88cm, height=2.45cm]{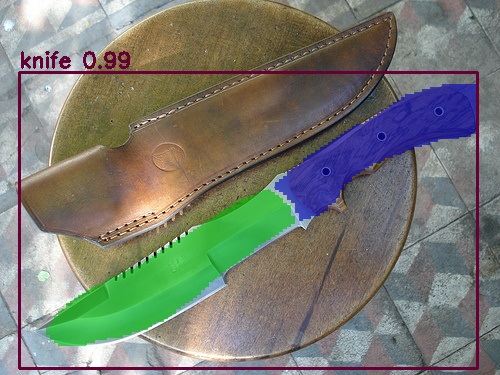}} {AffordanceNet112}
  \stackunder[7pt]{\includegraphics[width=2.88cm, height=2.45cm]{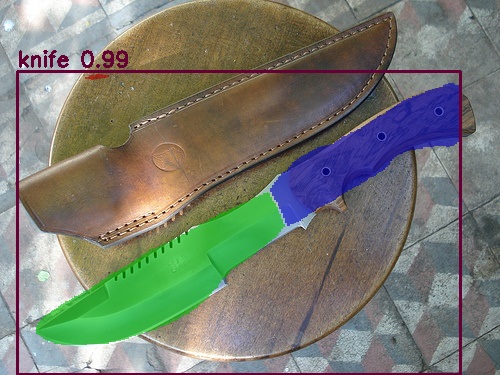}} {AffordanceNet244}
\vspace{3ex}
\caption{Examples of predicted affordance masks using different mask sizes. The predicted mask is smoother and finer when a bigger mask size is used.}

\label{Fig:result_mask_size_effect} 
\end{figure*}
\vspace{0.5ex}

\textbf{UMD Dataset} Table~\ref{tb_result_umd} summarizes results on the UMD dataset. On the average, our AffordanceNet also achieves the highest results on this dataset, i.e., it outperforms the second best (ED-RGBD) $2.9\%$. It is worth noting that the UMD dataset only contains clutter-free scenes, therefore the improvement of AffordanceNet over compared methods is not as high as the one in the real-world IIT-AFF dataset.
We recall that the AffordanceNet is trained using the RGB images only, while the second best (ED-RGBD) uses both RGB and the depth images. The Table~\ref{tb_result_umd} also clearly shows that the deep learning-based approaches such as AffordanceNet, DeepLab, ED-RGB significantly outperform the hand-designed geometric feature-based approaches (HMP, SRF).

To conclude, our AffordanceNet significantly improves over the state of the art, while it  does not require any extra post processing or data augmentation step. From the robotic point of view, AffordanceNet can be used in many tasks since it provides all the object locations, object categories, and object affordances in an end-to-end manner. The running time of AffordanceNet is around $150ms$ per image on a Titan X GPU, making it is suitable for robotic applications. Our implementation is based on Caffe deep learning library~\cite{DBLP:conf/mm/JiaSDKLGGD14}. The source code and trained models that allow reproducing the results in this paper will be released upon acceptance.

\subsection{Effect of Affordance Map Size}\label{Sec:exp_effect_mask_size}
\junk{The key difference between our AffordanceNet and other architectures for instance segmentation such as Mask-RCNN and FCIS is the use of the deconvolutional layer. In this section, we perform the following study to analyze the effect of the output affordance map size to the results:}
In this section, we analyze the effect of the affordance map size. Follow the setup in Mask-RCNN, we use only one deconvolutional layer with parameters ($d=1$, $s=2$, $S_f=4$) to create $14 \times 14$ affordance map from the $7 \times 7$ feature map (denoted as AffordanceNet14). Similarly, we change the parameters to ($d=1$, $s=4$, $S_f=6$) to create the $28 \times 28$ affordance map (denoted as AffordanceNet28). Furthermore, we also setup networks which use two deconvolutional layers to create $56 \times 56$ affordance map (denoted as AffordanceNet56), and three deconvolutional layers to create $112 \times 112$ affordance map (denoted as AffordanceNet112). Finally, to check the effect of the convolutional layers, we also setup a network with $6$ convolutional layers (together with ReLu), follow by a deconvolutional layer that upsampling the $7 \times 7$ feature map to $14 \times 14$ (denoted as AffordanceNet14\_6Conv).

\begin{figure}
  \centering
  
\subfigure[]{\label{fig_wild_a}\includegraphics[width=2.8cm, height=2.35cm]{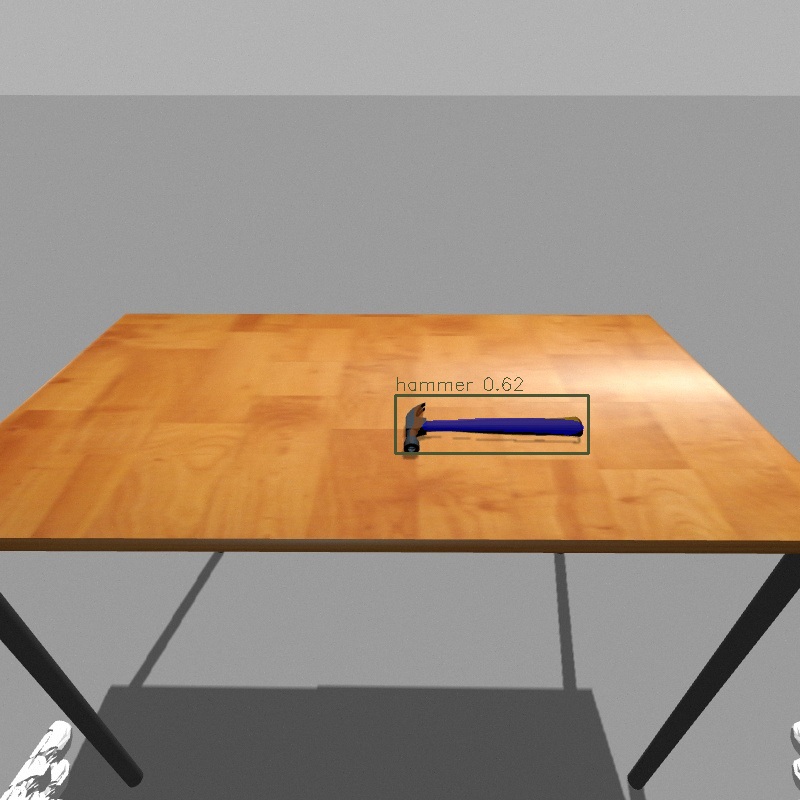}}
\subfigure[]{\label{fig_wild_b}\includegraphics[width=2.8cm, height=2.35cm]{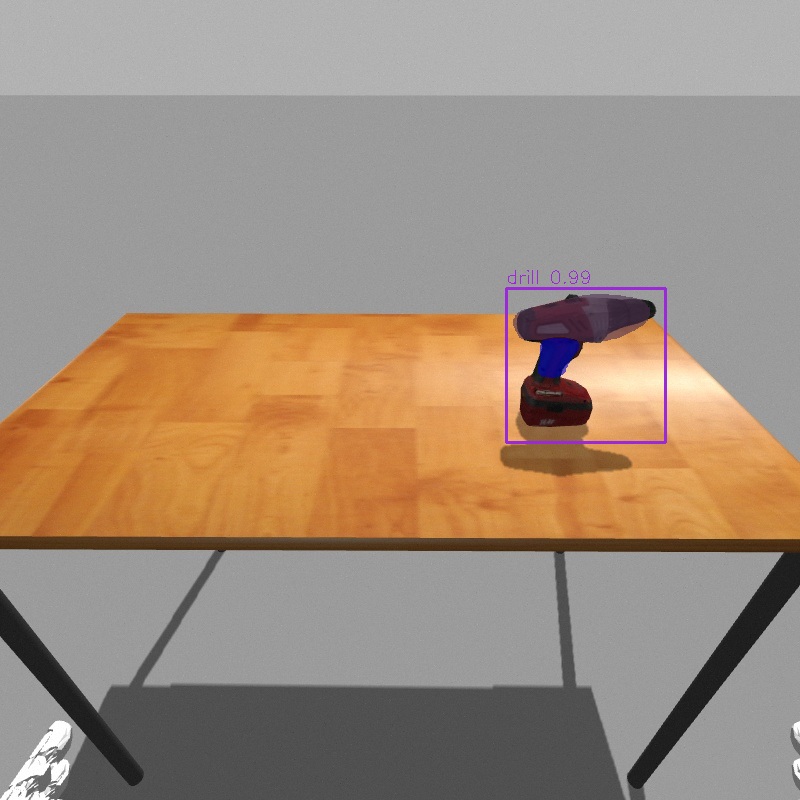}}
\subfigure[]{\label{fig_wild_c}\includegraphics[width=2.8cm, height=2.35cm]{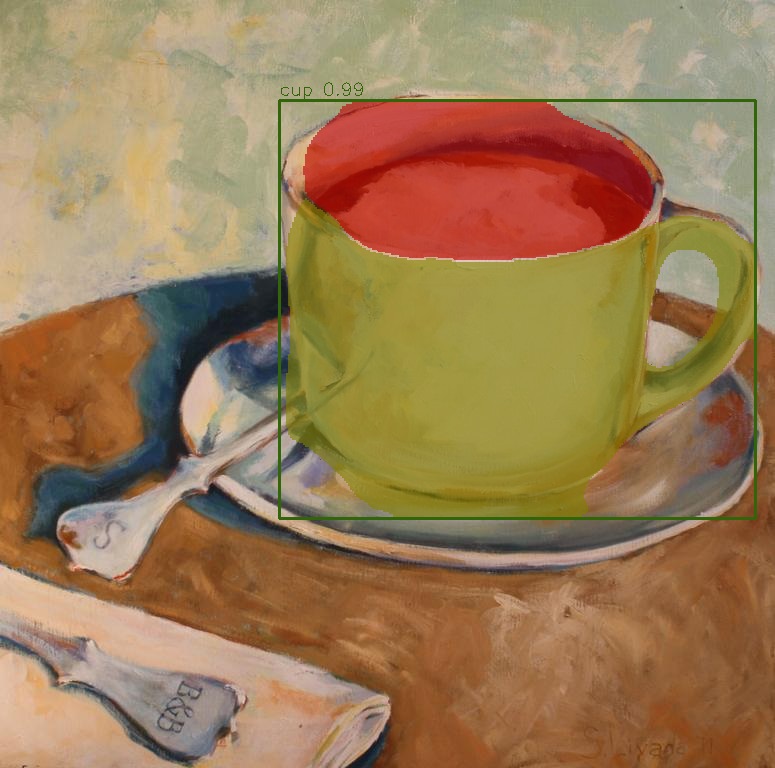}}

	\vspace{2ex}
    
 \caption{Affordance detection in the wild. \textbf{(a)} and \textbf{(b)}: We use AffordanceNet to detect the objects in Gazebo simulation. \textbf{(c)} AffordanceNet also performs well when the input is an artwork.}
 \label{Fig:result_aff_wild}
 \vspace{2ex}
\end{figure}

\vspace{0.3cm}
\begin{table}[!h]
\centering\ra{1.3}
\renewcommand\tabcolsep{8.0pt}
\caption{Effect of Mask Size}
\label{tb_result_masksize_effect}
\hspace{2ex}

\begin{tabular}{@{}rcccccccc@{}}
\toprule 					
& $F_\beta ^w$
\\
\midrule
AffordanceNet14 				& 57.71    \\
AffordanceNet28					& 66.13    \\
AffordanceNet56					& 71.54    \\
AffordanceNet112				& 72.52    \\ 
AffordanceNet14\_6Conv 			& 60.27    \\
\cline{1-2}	
AffordanceNet 					& 73.35    \\
 				
\bottomrule
\end{tabular}
\end{table}

Table~\ref{tb_result_masksize_effect} summarizes the average $F_\beta ^w$ score of the aforementioned networks on the IIT-AFF dataset. The results show that the affordance detection accuracy is gradually increasing when the bigger affordance map is used. In particular, the AffordanceNet14 gives very poor results since the map size of $14 \times 14$ is too small to represent multiclass affordances. The accuracy is significantly improved when we use the $28 \times 28$ affordance map. However, the improvement does not linearly increase with the affordance map size, it slows down when the bigger mask sizes are used. Note that using the big affordance map can improve the accuracy, but it also increases the number of parameters of the network. In our work, we choose the $244\times244$ map size for AffordanceNet since it both gives the good accuracy and can be trained with a Titan X GPU. We also found that using more convolutional layers (as in AffordanceNet14\_6Conv) can also improve the accuracy, but it still requires to upsample the affordance map to high resolution in order to achieve good results. Fig~\ref{Fig:result_mask_size_effect} shows some example results when different affordance map sizes are used.

\subsection{Affordance Detection in The Wild}

The experimental results on the simple constrained environment UMD dataset and  the real-world IIT-AFF dataset show that the AffordanceNet performs well on public research datasets. 
However, real-life images may be more challenging. In this study, we show some qualitative results to demonstrate that the AffordanceNet can generalize well in other testing environments. As illustrated in Fig~\ref{Fig:result_aff_wild}, 
 our AffordanceNet can successfully detect the objects and their affordances from artwork images or images from a simulated camera in Gazebo simulation~\cite{MingoHoffman2014}. 
Although this result is qualitative, it shows that AffordanceNet is applicable for wide ranges of applications, including in simulation environment which is crucial for developing robotic applications. 

\begin{figure}
\footnotesize
  \stackunder[5pt]{\includegraphics[width=2.79cm, height=3.18cm]{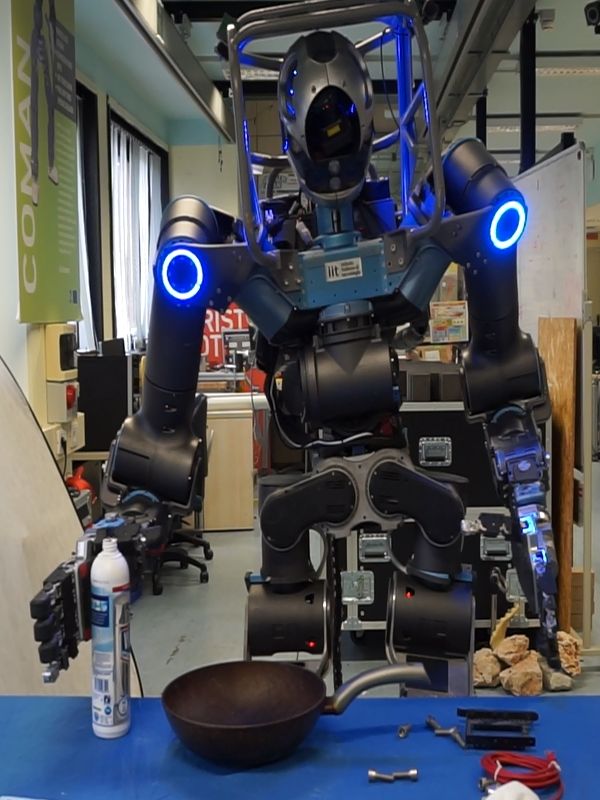}} {}\hspace{0.1cm}%
  \stackunder[5pt]{\includegraphics[width=2.79cm, height=3.18cm]{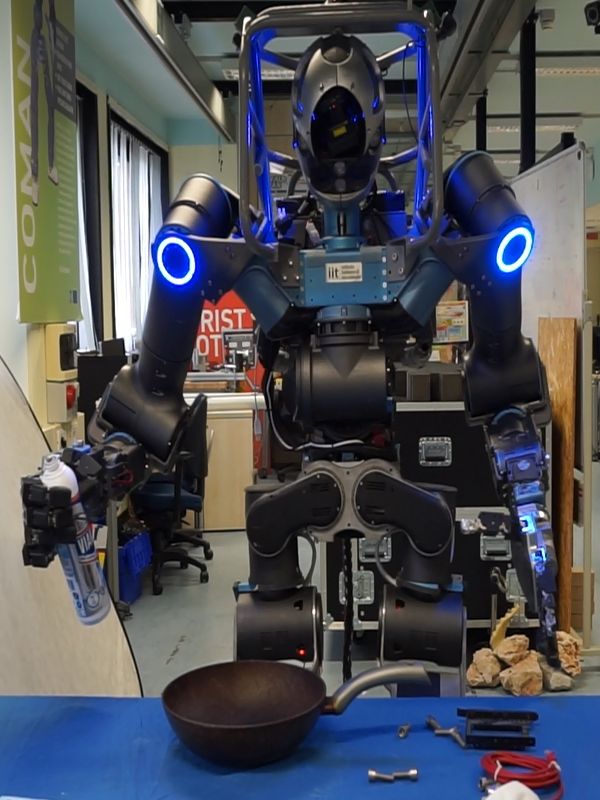}} {}\hspace{0.1cm}%
  \stackunder[5pt]{\includegraphics[width=2.79cm, height=3.18cm]{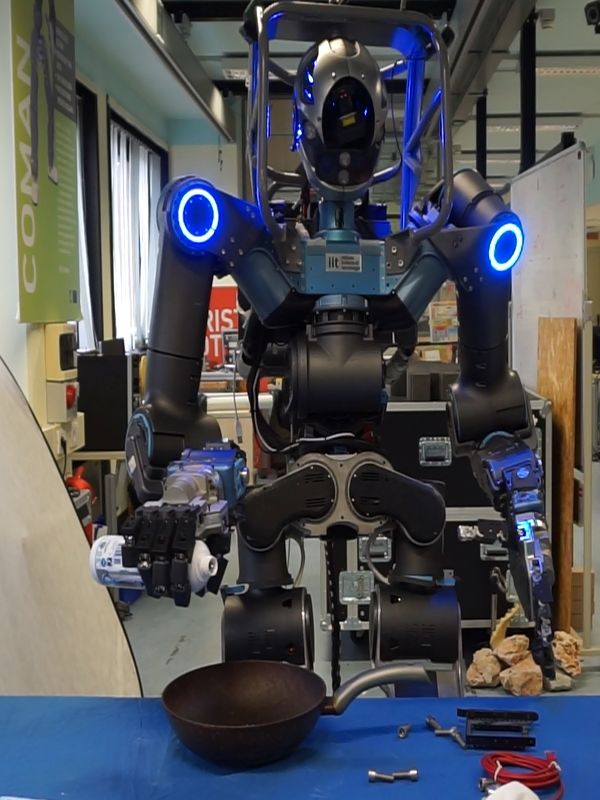}} {}
\vspace{3ex}
\caption{WALK-MAN is performing a pouring task. 
The outputs of the object detection branch help the robot to recognize and localize the objects (i.e., bottle, pan) while the outputs of the affordance detection branch help the robot to perform the task (i.e., where on the bottle the robot should grasp and where on the pan the water should be poured).}

\label{Fig:robot_pouring} 
\end{figure}

\subsection{Robotic Applications}
Since the AffordanceNet can detect both the objects and their affordances at the speed of $150ms$ per image, 
it is quite suitable for robotic applications. To demonstrate that, we use the humanoid robot WALK-MAN~\cite{Niko2016_full} to perform different manipulation experiments. The robot is controlled in real-time using the XBotCore framework~\cite{muratore2017xbotcore}. The whole-body motion planning is generated by OpenSoT library~\cite{Rocchi15}, while the AffordanceNet is used to provide visual information for the robot. Note that, from the 2D information outputted by AffordanceNet, we use the corresponding depth image to project it into 3D space, to be used in the real robot. Using this setup, the robot can perform different tasks such as grasping, pick-place, and pick-pouring. It is worth noting that all information produced by the AffordanceNet, i.e. the object locations, object labels, and object affordances are very useful for the tasks. 
For example, the robot knows where to grasp a \textsl{bottle} via the bottle's \texttt{grasp} affordance, and where to pour the water into a \textsl{pan} via the pan's \texttt{contain} affordance (see Fig.~\ref{Fig:robot_pouring}). Our experimental video can be found at the following link: {\url{https://sites.google.com/site/affordancenetwork/}}


%



\junk{
, no depth information is needed

}
\section{Conclusion}\label{Sec:con}
We have proposed AffordanceNet, an end-to-end deep learning framework that can simultaneously detect the objects and their affordances. Different from state-of-the-art network architectures for instance segmentation, we proposed three components to address the problem of multiple affordance classes in affordance detection task: a sequence of deconvolutional layers, a robust resizing strategy, and a new loss function. We showed that these components are the key factors to achieve high affordance detection accuracy. The extensive experimental results show that our AffordanceNet not only achieves state-of-the-art results on public datasets, but can also be used in various robotic applications.


\section*{Acknowledgment}
\addcontentsline{toc}{section}{Acknowledgment}
Thanh-Toan Do and Ian Reid are supported by the Australian Research Council through the Australian Centre for Robotic Vision (CE140100016). Ian Reid is also supported by an ARC Laureate Fellowship (FL130100102). Anh Nguyen is supported by the European Union Seventh Framework Programme (FP7-ICT-2013-10) under grant agreement no 611832 (WALK-MAN). The authors would like to thank Darwin G. Caldwell and Nikos G. Tsagarakis for the useful discussion.


\balance

\junk{
}

\bibliographystyle{class/IEEEtran}
\bibliography{class/clean_bib}
   
\end{document}